\documentclass[sigconf,nonacm]{acmart}

\AtBeginDocument{%
}

\setcopyright{none}
\renewcommand\footnotetextcopyrightpermission[1]{}

\usepackage{amsmath}
\usepackage{booktabs}
\usepackage{float}
\usepackage{xurl}
\graphicspath{{./}{figures/}}

\makeatletter
\newcommand{\fixedsubsection}[1]{%
  \@startsection{subsection}{2}{\z@}%
    {-.75\baselineskip}%
    {.25\baselineskip}%
    {\ACM@NRadjust\@subsecfont}{#1}}
\renewcommand{\@fnsymbol}[1]{%
  \ifcase#1\or\textdagger\or *\or\textdaggerdbl\or\S\or\P
  \or\ensuremath{\|}\or **\or\textdagger\textdagger
  \or\textdaggerdbl\textdaggerdbl\else\@ctrerr\fi}
\makeatother

\begin{document}

\title[PRIME for Shared CTR Top Networks]{PRIME: Mitigating Subgroup Optimization Competition in Shared CTR Top Networks with Plug-in Residual Input-Conditioned Mixture of Experts}

\author{Heng Yao}
\authornote{Corresponding authors: Heng Yao (\href{mailto:yaoheng.cs@qq.com}{yaoheng.cs@qq.com}) and Siyun Hou (\href{mailto:housiyun@hpu.edu.cn}{housiyun@hpu.edu.cn}).}
\affiliation{%
  \institution{Ant Group}
  \country{}
}

\author{Siyun Hou}
\authornotemark[1]
\affiliation{%
  \institution{Henan Polytechnic University}
  \country{}
}

\author{Tianying Liu}
\affiliation{%
  \institution{Independent Researcher}
  \country{}
}

\author{Yulou Shu}
\affiliation{%
  \institution{Alibaba Inc.}
  \country{}
}

\author{Yong He, Chuan Yuan, Kaibin Qiu, Guowei Chen, Jiayu Zhao, Chao Yu, and Ke Ding}
\affiliation{%
  \institution{Ant Group}
  \country{}
}

\begin{abstract}
Click-through rate (CTR) models vary in feature-interaction design, yet their top networks usually remain a single multilayer perceptron shared by all examples. Heterogeneous user, item, and context subgroups therefore update the same parameters; weakly aligned learning signals make the aggregate gradient a compromise among competing directions. We study the competition on Avazu with 4 models and 4 semantic fields. Across all architectures, semantic subgroups show lower Top-NN gradient cosine similarity than random groups matched by sample size and label ratio, with reductions of 0.23--0.37. This gap shows that semantic heterogeneity imposes less compatible optimization demands on a shared Top-NN.

This competition motivates input-conditioned experts, but directly replacing an established Dense mapping changes its initial function, sharing pattern, and capacity, obscuring the source of gains. We introduce PRIME (Plug-in Residual Input-conditioned Mixture of Experts), a Dense-anchored mixture of low-rank residual experts. PRIME anchors the original prediction and uses zero-residual initialization to match the Dense baseline exactly at training onset. Input-dependent routing weights low-rank experts for example-specific logit corrections; multi-bag aggregation and EMA load biases stabilize conditional estimation.

We evaluate PRIME on held-out Avazu and Criteo test sets across 13 CTR architectures and five paired seeds. PRIME improves mean AUC for 11 of 13 architectures on each dataset. Median paired AUC gains are +0.0022 and +0.0066, with LogLoss reductions of 0.0011 and 0.0081, respectively. On FiBiNET and DCNv2, PRIME outperforms APG in all ten seed-level AUC comparisons while using fewer parameters and lower inference latency on both backbones. These results show that function-preserving conditional residuals add input-dependent capacity while preserving the Dense path and its optimization stability. Code is available at \href{https://github.com/YH-learning/PRIME}{\textcolor{blue}{\nolinkurl{https://github.com/YH-learning/PRIME}}}.
\end{abstract}




\maketitle
\vspace{-0.6\baselineskip}

\section{Introduction}

\label{sec:introduction}

\vspace{-0.25\baselineskip}

\subsection{CTR Prediction and the Fully Shared Top-NN}

\label{sec:ctr-top-networks}

Click-through rate (CTR) prediction is a basic component of recommendation ranking and online advertising. Given user, item, and context features, a CTR model estimates the probability that an impression will receive a click. This probability affects candidate ranking, traffic allocation, ad auctions, and downstream value estimation. A useful CTR model must therefore balance ranking discrimination, probability quality, training stability, and inference cost \cite{ref04,ref23}.\footnote{BARS-CTR reports that, in large-scale production CTR systems, an absolute AUC or LogLoss improvement of approximately 0.001 is often considered practically meaningful \cite{ref04}. This value serves as an industrial rule of thumb for practical effect magnitude, distinct from a statistical-significance threshold.}

A typical CTR model comprises an embedding layer, a feature-interaction module, and a top prediction network. Sparse categorical features are first mapped to dense vectors. An interaction module then combines information across fields, and a shared decision path produces the click probability. We use \emph{Top-NN} to denote the shared MLP decision path found in most deep CTR architectures. Architectural innovation over the past decade has concentrated largely on the interaction module. The DCN family constructs explicit feature crosses \cite{ref18,ref19}, AutoInt learns field relations through self-attention \cite{ref15}, FiBiNET introduces bilinear interactions \cite{ref08}, and MaskNet modulates features with instance-specific masks \cite{ref20}. Despite their different inductive biases, these models typically retain a Dense Top-NN whose parameters are updated by every training example.

This design makes a consequential but rarely tested assumption: examples may require different interaction mechanisms, but a single post-interaction mapping is sufficient for all of them. A shared MLP is parameter-efficient and expressive, which has made it a reliable default. CTR data, however, are heterogeneous along user, advertisement, site, device, and temporal dimensions. Whether one fully shared top mapping remains the right parameterization under such heterogeneity has received little direct examination.

Semantic subgroups may rely on different predictive cues. The same advertisement can elicit different responses across mobile and desktop traffic, while in-app and web traffic may depend on different feature combinations. A shared Top-NN represents these patterns with the same parameters. Sharing pools observations but provides no mechanism for selecting input-specific parameter directions. The resulting constraint concerns optimization, not capacity alone. If subgroup gradients align, an aggregate update can improve several groups; as alignment decreases, the optimizer must compromise among subgroup objectives. We call this \emph{subgroup gradient competition}. Widening the MLP increases capacity but leaves every added unit active for every example, so it does not alter the underlying sharing constraint.

Ordinary scaling asks how many parameters are available; we ask whether different examples should use different parameters. If observable subgroup competition can be reduced by input-conditioned parameters, MoE becomes a way to reorganize sharing rather than merely enlarge the model. Section~\ref{sec:problem-formulation} formalizes this distinction.

\subsection{Diagnosing Subgroup Gradient Competition}

\label{sec:diagnosis-intro}

We diagnose this constraint on trained Dense models from Avazu. Model parameters are frozen, and semantic subgroups are constructed from \texttt{site\_category}, \texttt{app\_category}, \texttt{device\_type}, and \texttt{hour}. For each subgroup, we compute the loss gradient with respect to the shared Top-NN parameters. Gradient cosine similarity measures agreement between the corresponding update directions. Each semantic grouping is compared against random groups with the same sample size and positive-label ratio, controlling for two basic sources of gradient variation.

\begin{figure}[t]
  \centering
  \includegraphics[width=0.82\columnwidth]{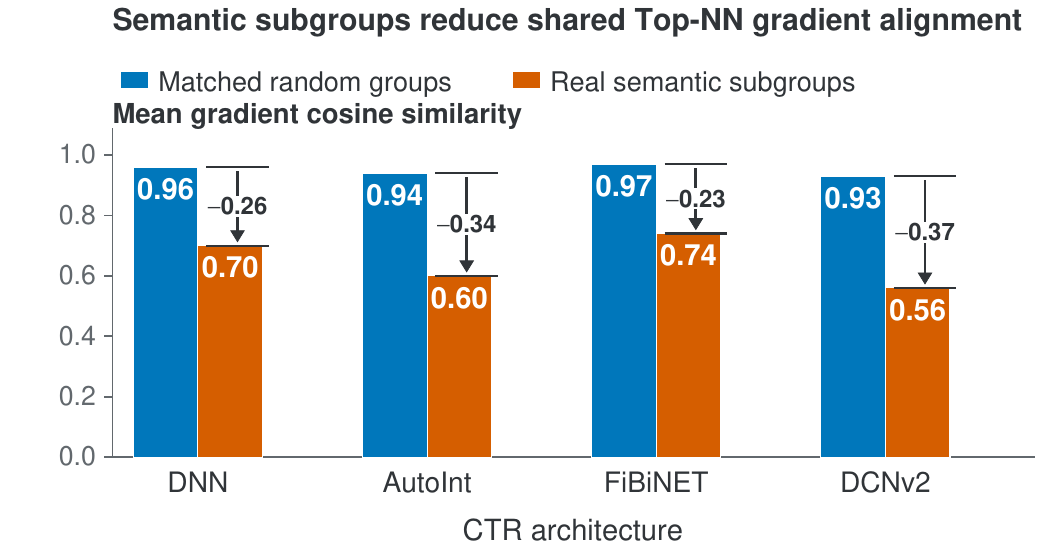}
  \caption{Semantic subgroups exhibit lower gradient alignment in the shared Top-NN. Each bar averages three trained checkpoints and four semantic fields. Blue bars denote matched random groups, and orange bars denote semantic subgroups. The short horizontal markers and arrows show the difference between them. Semantic subgroups have lower gradient similarity in all four architectures.}
  \Description{Gradient similarity in the shared Top-NN for semantic subgroups and matched random groups on Avazu. The chart or architecture is interpreted quantitatively in the caption and surrounding text.}
  \label{fig:prime-1}
  \vspace{-0.9\baselineskip}
\end{figure}

As Figure~\ref{fig:prime-1} shows, semantic partitioning consistently reduces gradient alignment relative to the matched random control: the gaps are 0.26 for DNN, 0.34 for AutoInt, 0.23 for FiBiNET, and 0.37 for DCNv2. Because each control preserves subgroup size and label prevalence, these substantial gaps cannot be explained by either factor. They instead reveal a systematic optimization effect: real semantic groups impose less compatible update directions on the shared Top-NN than composition-matched random groups. We refer to this excess loss of alignment as the \emph{subgroup competition gap}. Section~\ref{sec:gradient-diagnosis} formalizes the diagnostic and tests whether PRIME narrows this gap by providing input-conditioned residual directions.

\subsection{From Diagnosis to Function-Preserving Conditional Adaptation}

\label{sec:conditional-adaptation}

Mixture-of-Experts models change the granularity of parameter sharing by routing examples to local experts \cite{ref09,ref14,ref16}. Modern routing improves expert organization and stability \cite{ref34,ref35,ref36,ref42,ref43}, but transferring this conditional parameterization to CTR is not straightforward.

CTR prediction differs from language modeling in several respects. Each example usually supplies only one binary label; predicted probabilities affect both ranking and calibration; and the backbone already contains an explicit, architecture-specific interaction module. Adding MoE to a CTR Top-NN must therefore address three linked questions. The method should protect a strong Dense function, separate conditional specialization from ordinary capacity growth, and avoid duplicating interaction patterns already modeled by the backbone.

These requirements determine PRIME's design. New parameters start from the original Dense function, isolating the value of conditional adaptation from changes to the baseline mapping. Expert weighting depends explicitly on the input and delivers gains beyond parameter- and compute-matched controls. Each expert produces a compact low-rank logit correction instead of replicating the interaction module. Multi-bag aggregation combines several conditional estimates, while auxiliary-loss-free load adjustment maintains effective expert utilization without changing the CTR objective. Together, these mechanisms provide a stable operating layer around PRIME's function-preserving conditional parameterization.

Our central claim is that the limitation of a shared Top-NN lies in the granularity of parameter sharing, not in an inherent lack of MLP expressiveness. An input-conditioned low-rank residual can reduce the optimization compromise among heterogeneous examples while retaining the common Dense mapping.

This paper asks:

\begin{center}

\setlength{\fboxsep}{5pt}

\setlength{\fboxrule}{0.55pt}

\fbox{\parbox{0.90\columnwidth}{\centering\itshape How can a shared CTR Top-NN acquire input-conditioned correction directions without changing the backbone interaction module, disturbing the initial Dense function, or incurring uncontrolled computation?\par}}

\end{center}

The question leads to three primary requirements: function preservation, input-conditioned specialization, and budget-controlled attachment across backbones. Multi-bag aggregation and load adjustment are supporting implementation choices. Section 3 derives PRIME from this hierarchy after formalizing subgroup optimization under a shared Top-NN.

The resulting contributions are threefold:

\begin{enumerate}
  \item To the best of our knowledge, we present the first controlled empirical study of semantic subgroup competition in the shared Top-NN of single-task CTR models. The diagnostic compares real semantic partitions with random partitions matched for size and label ratio, and verifies the competition gap across CTR architectures and seeds.
  \item We propose PRIME, a function-preserving conditional residual expert model. PRIME retains the shared Dense mapping as an anchor and uses input-dependent routing to combine low-rank logit corrections, separating conditional capacity from the backbone's feature-interaction mechanism.
  \item Across 13 CTR architectures, two datasets, and five paired seeds, matched-budget controls, ablations, and gradient diagnostics separate input-conditioned organization from ordinary capacity expansion.
\end{enumerate}

\section{Related Work}

\label{sec:related-work}

\subsection{CTR Feature Interaction and Shared Prediction Layers}

\label{sec:related-ctr}

CTR modeling has long focused on interactions among sparse features. Factorization Machines represent second-order interactions with low-rank inner products \cite{ref13}. Wide\&Deep and DeepFM combine low-order and deep nonlinear paths \cite{ref01,ref07}, while NFM, PNN, AFM, and FwFM refine interaction pooling, product layers, attention, and field weighting, respectively \cite{ref06,ref11,ref12,ref22}. DCN/DCNv2, xDeepFM, AutoInt, FiBiNET, and MaskNet build richer representations through explicit crosses, compressed interactions, self-attention, bilinear interactions, and instance-guided masks \cite{ref08,ref15,ref18,ref19,ref20,ref26}. DIN activates historical interests relative to a target item \cite{ref27}, and FinalMLP constructs complementary MLP streams through stream-specific feature selection \cite{ref28}. These methods alter the input representation or interaction path, but their final decisions are usually produced by MLP parameters shared across all examples. PRIME leaves the interaction module unchanged and instead modifies how Top-NN capacity is shared.

Viewed from the decision layer, these architectures differ in how they transform the shared embedding $z$, yet converge on a globally shared prediction parameterization. PRIME targets this common endpoint rather than adding another interaction operator, allowing the same conditional adapter to be evaluated across otherwise heterogeneous backbones.

\vspace{-0.8\baselineskip}

\subsection{Conditional Parameterization and Expert Models for Recommendation}

\label{sec:related-conditional}

Conditional parameterization changes network weights as a function of the input or task. HyperNetworks generate the parameters of another network \cite{ref29}. APG dynamically produces low-rank prediction parameters for individual CTR examples \cite{ref30} and is one of the closest published methods to PRIME. STAR combines shared central parameters with domain-specific parameters \cite{ref31}, while PEPNet modulates embeddings and DNN units using user, domain, and task priors \cite{ref32}. More recent work has combined low-rank adaptation with expert selection. iLoRA uses sequence representations to gate LoRA experts for instance-wise adaptation in sequential recommendation \cite{ref37}. MLoRA assigns a low-rank adapter to each CTR domain and reports both offline and online gains \cite{ref38}. HiLoMoE uses hierarchical rank-one experts to scale CTR models in depth and width \cite{ref39}, whereas MoE-MLoRA first trains domain experts and then learns a gate for multi-domain CTR \cite{ref40}. MLoRA+ fuses domain-specific LoRA adapters with a Transformer to model cross-domain dependencies \cite{ref44}. These studies support low-rank experts as a compact representation of conditional variation, but they rely on pretrained language models, explicit domain boundaries, or multi-stage optimization. PRIME targets conventional single-task CTR without domain labels or staged training.

Expert architectures in recommendation have also been developed mainly for multi-task or multi-scenario sharing. MMoE combines shared experts through task-specific gates \cite{ref10}, and PLE separates shared from task-specific experts to reduce negative transfer \cite{ref17}. PRIME operates in one label space with one CTR objective. Latent semantic groups are used only for post-training diagnosis and never supervise the router. The problem is not expert allocation across observed tasks, but update competition among heterogeneous examples within a single task.

\vspace{-0.65\baselineskip}

\subsection{MoE Stability and Controlled Residual Adaptation}

\label{sec:related-moe}

Classical MoE learns local specialization through gating \cite{ref09}, and sparse gating and Switch Transformers use conditional activation to scale model capacity \cite{ref05,ref14}. Routing instability, expert redundancy, and load imbalance have consequently become central concerns. ST-MoE studies stable training for sparse expert models \cite{ref16}, Expert Choice Routing assigns a fixed capacity to each expert \cite{ref33}, Soft MoE avoids token dropping and discrete assignment through continuous soft routing \cite{ref34}, DeepSeekMoE increases specialization with fine-grained routed and shared experts \cite{ref35}, and ReMoE replaces discontinuous Top-$k$ selection with ReLU routing \cite{ref36}. More recent work addresses whether routes reflect expert capabilities: ERC loss explicitly couples routers and experts \cite{ref42}, while RoMA aligns routing-weight and task-semantic manifolds to improve generalization \cite{ref43}. Auxiliary-loss-free balancing instead adjusts dynamic expert biases without adding balancing gradients to the task objective \cite{ref21}. These methods primarily address sparse computation at language-model scale. PRIME uses a small low-rank soft-routed module and concentrates on function preservation, controlled conditional estimation, and adaptation across CTR architectures.

PRIME also draws on low-rank adaptation, residual learning, and ensemble estimation. LoRA confines parameter updates to a low-dimensional subspace \cite{ref02}, and iLoRA, MLoRA, and HiLoMoE extend low-rank parameterization to instance-, domain-, and hierarchy-conditioned recommendation \cite{ref37,ref38,ref39}. Residual learning preserves a baseline mapping \cite{ref03}, while bagging reduces variance by averaging diverse predictors \cite{ref24}. PRIME combines these ideas in a single end-to-end optimization: low rank restricts expert freedom, zero-residual initialization preserves the initial Dense function, and averaging smooths multiple conditional estimates. Unlike a fixed global low-rank update, a staged expert system, or independently trained full predictors, PRIME routes residual directions by input while sharing the same CTR backbone.

Existing work studies task-level negative transfer and domain-conditioned parameters. PRIME instead identifies semantic subgroup competition within the shared Top-NN of a single CTR task. Controls matched for group size and click rate separate this gap from composition effects, motivating a conditional residual design that preserves the Dense function.

\vspace{-0.8\baselineskip}

\section{PRIME}

\label{sec:method}

\begin{figure}[t]
  \centering
  \includegraphics[width=0.98\columnwidth]{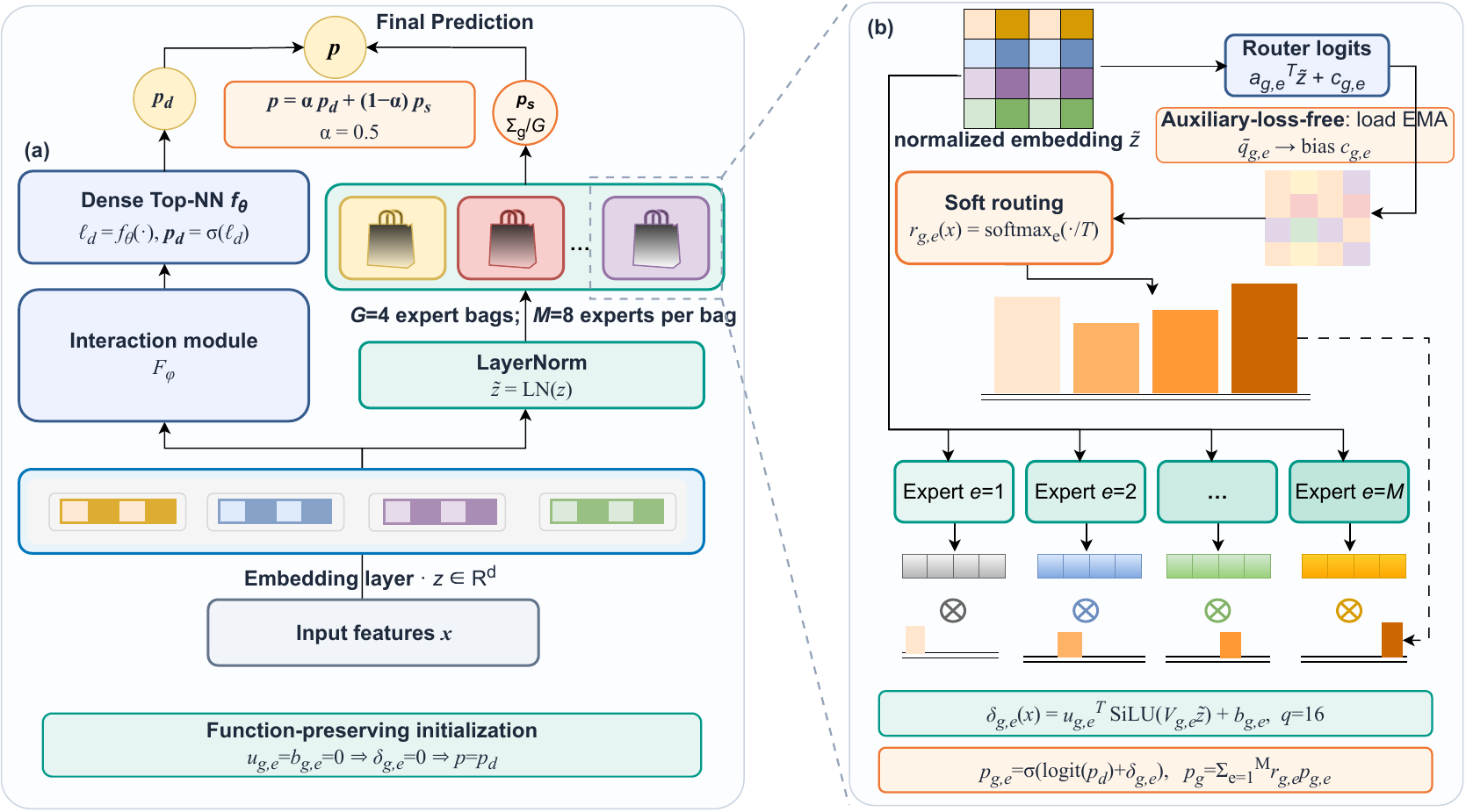}
  \caption{PRIME architecture. The original model produces $p_d$; $G=4$ bags route $z$ to rank-$q$ logit-residual experts and average their predictions into $p_s$. The final prediction mixes $p_d$ and $p_s$ with $\alpha=0.5$. EMA load biases adjust utilization without auxiliary gradients, and zero-residual initialization gives $p=p_d$ at the start of training.}
  \Description{PRIME architecture with bag-level routing and low-rank residual experts. The chart or architecture is interpreted quantitatively in the caption and surrounding text.}
  \label{fig:prime-2}
  \vspace{-0.8\baselineskip}
\end{figure}

Figure~\ref{fig:prime-2} summarizes PRIME's two-path architecture. The original CTR backbone produces the anchored probability $p_d$, while $G$ routed low-rank expert bags estimate the conditional probability $p_s$; their convex combination yields the final prediction $p$.

\subsection{Problem Formulation and Shared Optimization}

\label{sec:problem-formulation}

Let $x$ denote an input example. The embedding layer $E_{\omega}$ produces the flattened representation $z$, and the complete original CTR backbone $B_{\phi,\theta}$ produces the baseline logit $\ell_d$ and probability $p_d$:

\begin{equation}
\label{eq:1}
z=\operatorname{vec}(E_{\omega}(x)),\qquad
\ell_d=B_{\phi,\theta}(z),\qquad
p_d=\sigma(\ell_d).
\end{equation}

Here $\phi$ denotes the interaction parameters and $\theta$ the shared Top-NN parameters. When a backbone contains parallel shallow or interaction logits, $B_{\phi,\theta}$ includes those paths as well, so $p_d$ is always the complete original prediction. PRIME leaves every component of $B_{\phi,\theta}$ unchanged and attaches its residual branch to $z$ and $p_d$.

Suppose the training distribution contains $K$ latent subgroups. Subgroup $k$ has probability mass $\pi_k$ and loss $\mathcal L_k(\theta)$. The objective and gradient of the shared Top-NN are

\begin{equation}
\label{eq:2}
\mathcal L(\theta)=\sum_{k=1}^{K}\pi_k\mathcal L_k(\theta),
\qquad
g=\nabla_{\theta}\mathcal L(\theta)=\sum_{k=1}^{K}\pi_kg_k,
\end{equation}

where $g_k=\nabla_{\theta}\mathcal L_k(\theta)$. For an update $\theta'=\theta-\eta g$, the first-order change in subgroup $k$'s loss is

\begin{equation}
\label{eq:3}
\scalebox{0.92}{${\displaystyle \begin{aligned}
\mathcal L_k(\theta')-\mathcal L_k(\theta)
&\approx -\eta g_k^{\top}g \\
&=-\eta\left[
\pi_k\lVert g_k\rVert_2^2+
\sum_{j\neq k}\pi_j\lVert g_k\rVert_2\lVert g_j\rVert_2
\cos(g_k,g_j)
\right].
\end{aligned}}$}
\end{equation}

Equation~\ref{eq:3} shows that the effective descent a subgroup receives from a shared update depends on its gradient alignment with the other subgroups. Lower cosine similarity weakens the aggregate update for that subgroup. If the similarity becomes negative, an update that benefits one subgroup can increase another subgroup's loss. A fully shared Top-NN must therefore compromise among subgroup objectives.

Let $\theta_k^{\star}$ be the optimum obtained by training only on subgroup $k$, and let $\theta^{\star}$ be the optimum under fully shared parameters. The compromise due to sharing can be written as

\begin{equation}
\label{eq:4}
\mathcal G_{\mathrm{share}}
=\sum_{k=1}^{K}\pi_k
\left[\mathcal L_k(\theta^{\star})-
\mathcal L_k(\theta_k^{\star})\right]\ge 0.
\end{equation}

Equation~\ref{eq:4} motivates a compact intervention: input-dependent low-dimensional corrections reduce the sharing compromise while retaining the statistical efficiency of a common baseline. PRIME therefore preserves the complete baseline function and introduces a conditional adapter $A_{\psi}$:

\begin{equation}
\label{eq:5}
p=A_{\psi}(z,p_d).
\end{equation}

The adapter must satisfy three primary conditions.

\textbf{Function preservation.} At initialization $\psi_0$,

\begin{equation}
\label{eq:6}
A_{\psi_0}(z,p_d)=p_d,\qquad \forall z.
\end{equation}

\textbf{Input-conditioned parameter organization.} For two inputs $x_i$ and $x_j$, the routing weights may satisfy $r(x_i)\neq r(x_j)$, preventing the added capacity from collapsing into a fixed ensemble. Conditional organization should retain a measurable advantage when parameter count or computation is matched.

\textbf{Budget control and backbone decoupling.} The added parameters and operations should be governed by a low rank $q\ll d$. Experts output generic logit residuals rather than replicating a complete Top-NN or the backbone's interaction module.

PRIME adds two supporting implementation choices: multiple expert bags aggregate alternative conditional estimates, and a gradient-free load bias limits extreme routing imbalance. Both are trained within the same end-to-end binary cross-entropy objective. Ablations distinguish these refinements from the three conditions that define PRIME's core parameterization.

\subsection{Dense-Anchored Conditional Residual Parameterization}

\label{sec:dense-anchor}

For latent subgroup $k$, write its ideal logit as a shared baseline plus a subgroup-specific correction:

\begin{equation}
\label{eq:7}
\ell_k^{\star}(x)=\ell_d(x)+\Delta_k(x).
\end{equation}

This decomposition assigns common predictive structure to $\ell_d$ and focuses each expert on a residual $\Delta_k$. Following the baseline-preserving principle of residual learning \cite{ref03}, the new branch learns conditional deviations around an established prediction function. Since subgroup labels are unavailable at inference time, PRIME uses input-dependent routing to weight candidate residuals:

\begin{equation}
\label{eq:8}
\Delta(x)\approx\sum_e r_e(x)\delta_e(x).
\end{equation}

The Dense path is a functional anchor shared by all experts, while the routed branch learns conditional corrections around the original model function. A fixed mixture weight bounds the expert contribution and guarantees a persistent Dense component throughout training.

\subsection{Low-Rank Residual Experts}

\label{sec:low-rank-experts}

PRIME obtains a flattened embedding representation $z\in\mathbb R^d$ from the original model and applies Layer Normalization without affine parameters:

\begin{equation}
\label{eq:9}
\tilde z=\operatorname{LN}(z).
\end{equation}

Expert $e$ in bag $g$ is defined as

\begin{equation}
\label{eq:10}
\delta_{g,e}(x)
=u_{g,e}^{\top}
\operatorname{SiLU}\!\left(V_{g,e}\tilde z\right)+b_{g,e},
\end{equation}

where $V_{g,e}\in\mathbb R^{q\times d}$, $u_{g,e}\in\mathbb R^q$, and $q\ll d$. Matrix $V_{g,e}$ projects the input into a $q$-dimensional subspace, after which $u_{g,e}$ maps the nonlinear response to a scalar logit correction. As in low-rank adaptation \cite{ref02}, this restriction limits each expert to $O(qd)$ parameters, far fewer than a replicated Top-NN. Distinct matrices $V_{g,e}$ represent different low-dimensional correction directions, which the router combines according to the input.

To satisfy the function-preservation condition in Equation~\ref{eq:6}, PRIME initializes $u_{g,e}$ and $b_{g,e}$ to zero:

\begin{equation}
\label{eq:11}
\delta_{g,e}(x)=0,\qquad \forall g,e,x.
\end{equation}

The projection matrices and routers may start from different directions without changing the initial prediction. Optimization first develops the expert output projections and only then establishes router-expert specialization, preventing untrained experts from dominating early predictions.

\subsection{Input-Conditioned Routing, Bag Aggregation, and Functional Equivalence}

\label{sec:routing-bagging}

Each expert bag has an independent router. The routing probability within bag $g$ is

\begin{equation}
\label{eq:12}
r_{g,e}(x)=
\frac{\exp\left((a_{g,e}^{\top}\tilde z+c_{g,e})/T\right)}
{\sum_{j=1}^{M}\exp\left((a_{g,j}^{\top}\tilde z+c_{g,j})/T\right)},
\end{equation}

where $M$ is the number of experts per bag, $T$ is the routing temperature, and $c_{g,e}$ is a load-correction bias. The main experiments use $T=1$. Each expert corrects the baseline logit:

\begin{equation}
\label{eq:13}
p_{g,e}(x)=
\sigma\!\left(\operatorname{logit}(p_d(x))+\delta_{g,e}(x)\right).
\end{equation}

The bag prediction is the conditional expectation of the expert probabilities:

\begin{equation}
\label{eq:14}
p_g(x)=\sum_{e=1}^{M}r_{g,e}(x)p_{g,e}(x).
\end{equation}

Equations~\ref{eq:12} and~\ref{eq:14} make the expert mixture input-dependent. Replacing $r_{g,e}(x)$ with a constant turns the branch into an unconditional ensemble. Permuting $r(x)$ across examples preserves aggregate load while breaking the association between an input and its expert weights. Section~\ref{sec:capacity-controls} uses both interventions to test the contribution of conditional routing.

\textbf{Bag aggregation.} PRIME partitions $E=GM$ experts into $G$ bags to obtain multiple conditional estimates under a fixed total expert count. Each bag independently evaluates Equations~\ref{eq:12}--\ref{eq:14}, and the bag predictions are averaged:

\begin{equation}
\label{eq:15}
p_s(x)=\frac{1}{G}\sum_{g=1}^{G}p_g(x).
\end{equation}

If bag predictions have variance $\sigma_p^2$ and an approximate pairwise correlation $\rho_p$, the variance of their mean is

\begin{equation}
\label{eq:16}
\operatorname{Var}[p_s]
\approx\frac{\sigma_p^2}{G}
\left[1+(G-1)\rho_p\right].
\end{equation}

Equation~\ref{eq:16} motivates prediction averaging as an output-smoothing mechanism under correlated conditional estimates. The bags share a backbone and are optimized jointly, so the expression motivates the design without assuming independent bagging estimators. The single-bag comparison in Section~\ref{sec:capacity-controls} shows that multi-bag aggregation contributes complementary output smoothing, with gains concentrated on backbones that benefit most from routing diversity.

\textbf{Dense probability anchor.} The final prediction is a convex combination of the Dense and expert probabilities:

\begin{equation}
\label{eq:17}
p(x)=\alpha p_d(x)+(1-\alpha)p_s(x),
\qquad 0\le \alpha\le 1.
\end{equation}

We use $\alpha=0.5$. Equation~\ref{eq:17} bounds expert corrections in probability space and retains an explicit Dense contribution throughout training. The original path remains in the computation graph even if routing becomes highly concentrated.

From Equation~\ref{eq:11},

\begin{equation*}
p_{g,e}(x)=\sigma(\operatorname{logit}(p_d(x)))=p_d(x).
\end{equation*}

Substitution into Equations~\ref{eq:14}--\ref{eq:17} gives

\begin{equation}
\label{eq:18}
p_g(x)=p_s(x)=p(x)=p_d(x).
\end{equation}

PRIME is therefore exactly function-equivalent to the Dense model at initialization. This property follows directly from zero-residual initialization and convex mixing, without relying on hyperparameter tuning.

\subsection{Auxiliary-Loss-Free Load Adjustment}

\label{sec:load-adjustment}

Sparse MoE models often add a load-balancing term to the task loss to prevent a small number of experts from absorbing most inputs \cite{ref14}:

\begin{equation}
\label{eq:19}
\mathcal L_{\mathrm{total}}
=\mathcal L_{\mathrm{CTR}}+\lambda_{\mathrm{bal}}\mathcal L_{\mathrm{bal}}.
\end{equation}

The coefficient $\lambda_{\mathrm{bal}}$ creates an additional trade-off between expert utilization and the CTR objective. Following auxiliary-loss-free load balancing \cite{ref21}, PRIME does not backpropagate $\mathcal L_{\mathrm{bal}}$. It instead tracks average routing load within each bag. For batch $t$,

\begin{equation}
\label{eq:20}
\hat q_{g,e}^{(t)}=
\frac{1}{B}\sum_{i=1}^{B}r_{g,e}(x_i),
\end{equation}

and the exponential moving average is updated as

\begin{equation}
\label{eq:21}
\bar q_{g,e}^{(t)}
=\mu\bar q_{g,e}^{(t-1)}+(1-\mu)\hat q_{g,e}^{(t)}.
\end{equation}

The load bias is corrected without gradients toward the target load $1/M$:

\begin{equation}
\label{eq:22}
\begin{aligned}
\tilde c_{g,e}^{(t+1)}
&=c_{g,e}^{(t)}
+\eta_b\left(\frac{1}{M}-\bar q_{g,e}^{(t)}\right),\\
c_{g,e}^{(t+1)}
&=\operatorname{clip}\!\left(
\tilde c_{g,e}^{(t+1)}
-\frac{1}{M}\sum_{j=1}^{M}\tilde c_{g,j}^{(t+1)},
-c_{\max},c_{\max}
\right).
\end{aligned}
\end{equation}

Equation~\ref{eq:22} updates each bias from its load error and removes the shared offset by centering within a bag. We set $\mu=0.99$, $\eta_b=10^{-3}$, and $c_{\max}=2$. The update changes routing biases for subsequent batches but sends no additional gradient to model parameters. PRIME is consequently trained end to end with a single CTR objective:

\begin{equation}
\label{eq:23}
\scalebox{0.92}{${\displaystyle \mathcal L_{\mathrm{PRIME}}
=-\frac{1}{N}\sum_{i=1}^{N}
\left[y_i\log p_i+(1-y_i)\log(1-p_i)\right]
+\lambda_2\lVert\Theta\rVert_2^2.}$}
\end{equation}

\subsection{Parameter Count, Computation, and Plug-in Implementation}

\label{sec:complexity}

Let $d$ be the input dimension, $E$ the total number of experts, and $q$ the expert rank. Ignoring scalar activation costs, the routers require $Ed$ multiply-accumulate operations (MACs), and the low-rank experts require $Eq(d+1)$. The additional computation is

\begin{equation}
\label{eq:24}
C_{\mathrm{extra}}=Ed+Eq(d+1).
\end{equation}

The number of added trainable parameters is

\begin{equation}
\label{eq:25}
P_{\mathrm{extra}}
=Ed+Eqd+Eq+E
=E(q+1)(d+1).
\end{equation}

Both quantities are controlled by $q$, avoiding replication of $E$ complete Top-NNs.

At runtime, PRIME executes the original model to obtain $p_d$ and captures the flattened embedding representation $z$ through a hook, as shown in Figure~\ref{fig:prime-2}(a). It then evaluates Equations~\ref{eq:9}--\ref{eq:17}. Feature processing, the interaction module, the Dense Top-NN, the training entry point, and the record format remain unchanged. Enabling the plug-in reconstructs the optimizer so that the new parameters participate in the same end-to-end training process.

These design choices yield three primary predictions: matched non-conditional capacity should not reproduce PRIME's full gain, breaking input-route correspondence should reduce performance, and preserving the Dense anchor should improve robustness across backbones. The experiments test these predictions before the configuration is frozen for cross-architecture evaluation.

\section{Experiments}

\label{sec:experiments}

\subsection{Research Questions and Experimental Protocol}

\label{sec:protocol}

The experiments address five research questions in the order of the evidence presented.

\begin{itemize}
  \item \textbf{RQ1: Problem diagnosis (Section~\ref{sec:gradient-diagnosis}).} Do semantic subgroups exhibit lower gradient alignment in a shared Top-NN than matched random groups?
  \item \textbf{RQ2: Mechanism association (Section~\ref{sec:gradient-diagnosis}).} Does the semantic subgroup competition gap decrease after training with PRIME?
  \item \textbf{RQ3: Capacity and conditioning (Section~\ref{sec:capacity-controls}).} Can non-conditional Dense parameters or matched computation explain PRIME's gains, and is input-dependent routing necessary?
  \item \textbf{RQ4: Design hierarchy and efficiency (Sections 4.3--4.4).} Which PRIME components are necessary, and how does the resulting model compare with related conditional parameterization in accuracy and end-to-end cost?
  \item \textbf{RQ5: Cross-architecture and cross-dataset effectiveness (Section~\ref{sec:final-results}).} After freezing the configuration, how broadly does PRIME improve held-out performance across CTR interaction architectures on Avazu and Criteo?
\end{itemize}

We use the public FuxiCTR/BARS-CTR processing pipeline \cite{ref04,ref23}. Avazu contains 22 fields and 28.30M, 4.04M, and 8.09M examples in its training, validation, and test splits, respectively. Criteo contains 39 fields, with corresponding split sizes of 36.67M, 4.58M, and 4.58M. The evaluation covers 13 CTR architectures: AFM \cite{ref22}, AutoInt \cite{ref15}, DCN \cite{ref18}, DNN, DeepFM \cite{ref07}, FiBiNET \cite{ref08}, FwFM \cite{ref12}, GDCN \cite{ref25}, MaskNet \cite{ref20}, NFM \cite{ref06}, PNN \cite{ref11}, Wide\&Deep \cite{ref01}, and DCNv2 \cite{ref19}. For backbones without a standalone MLP head, \texttt{Dense} denotes the complete original prediction path; PRIME still anchors its unmodified probability $p_d$. Every Dense/PRIME pair shares its split, preprocessing, backbone configuration, and random seed. Model selection uses validation AUC, after which the untouched test set is evaluated over five paired seeds.\footnote{The paired seeds are 2021, 190034, 27011, 948432, and 992817.}

All main experiments use Adam with learning rate $10^{-3}$, batch size 4,096, and the public backbone hyperparameters. Training runs for at most 100 epochs, stops after two validation checks without improvement, and restores the best validation checkpoint. The embedding dimension is 10. PRIME is frozen at $G=4$ bags, $M=8$ experts per bag, rank $q=16$, Dense weight $\alpha=0.5$, and routing temperature $T=1$. Completed runs are retained; a job is restarted only if it exits abnormally or produces no valid record. OpenAI Codex was used to assist with experimental script development and code debugging; all generated code changes were reviewed and validated by the authors.

We define the paired metrics as

\begin{equation}
\label{eq:26}
\Delta\mathrm{AUC}
=\mathrm{AUC}_{\mathrm{PRIME}}-
\mathrm{AUC}_{\mathrm{Dense}},
\end{equation}

\begin{equation}
\label{eq:27}
\Delta\mathrm{LogLoss}
=\mathrm{LogLoss}_{\mathrm{Dense}}-
\mathrm{LogLoss}_{\mathrm{PRIME}}.
\end{equation}

Positive values indicate an improvement by PRIME for both definitions. We report the original metrics and five-seed means at the architecture level. %Avazu results additionally retain seed-level paired directions. No completed run is removed after observing its result.

\subsection{Diagnosing Shared Top-NN Competition and Its Change under PRIME}

\label{sec:gradient-diagnosis}

To answer RQ1, we evaluate DNN, AutoInt, FiBiNET, and DCNv2 with three trained checkpoints per architecture and four semantic fields. This gives 48 comparisons across architectures, checkpoints, and fields. Semantic subgroups are defined by \texttt{site\_\allowbreak category}, \texttt{app\_\allowbreak category}, \texttt{device\_\allowbreak type}, and \texttt{hour}. Each group is sampled into eight disjoint blocks containing 256 positive and 256 negative examples per block, for 4,096 examples in total. Matched random groups use exactly the same sample size and label ratio, with 2,000 random permutations. Models are placed in evaluation mode, and unregularized binary cross-entropy gradients are computed on the validation set. The gradient parameter set is restricted to all trainable weights in the shared Top-NN. Gradients are flattened before pairwise cosine similarities are calculated. Every architecture in Figure~\ref{fig:prime-1} satisfies $A_f^{\mathrm{sem}}<A_f^{\mathrm{rand}}$, establishing semantic grouping as a consistent source of lower agreement among shared updates after controlling for group size and label ratio.

For semantic field $f$, let $A_f^{\mathrm{sem}}$ be the mean pairwise gradient similarity among semantic subgroups and $A_f^{\mathrm{rand}}$ that of matched random groups. Define the competition gap as

\begin{equation}
\label{eq:28}
\Gamma_f=A_f^{\mathrm{rand}}-A_f^{\mathrm{sem}}.
\end{equation}

A larger $\Gamma_f$ indicates greater gradient heterogeneity associated with semantic grouping. To address RQ2, we compare trained Dense and PRIME models using the same architectures, seeds, fields, and examples.

\begin{figure}[t]
  \centering
  \includegraphics[width=\columnwidth]{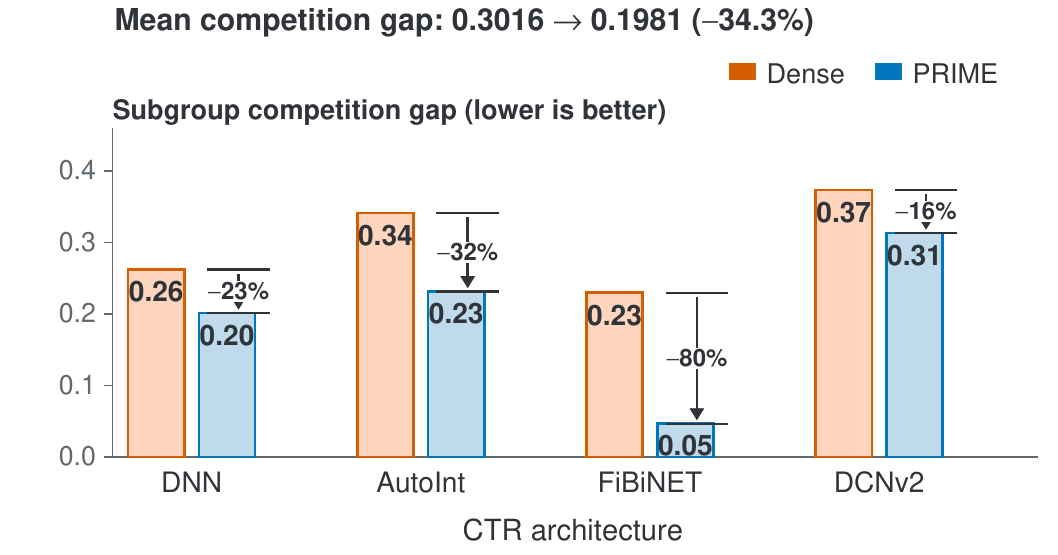}
  \caption{PRIME reduces the competition gap between semantic and matched random groups. Bars report architecture-level means over three paired checkpoints and four semantic fields. Percentages denote the relative reduction from Dense to PRIME; lower values indicate more consistent subgroup updates.}
  \Description{PRIME reduces subgroup gradient competition in the shared Top-NN across four CTR architectures. The chart or architecture is interpreted quantitatively in the caption and surrounding text.}
  \label{fig:prime-3}
\end{figure}

As shown in Figure~\ref{fig:prime-3}, the mean competition gap across the four architectures decreases from $\bar\Gamma_{\mathrm{Dense}}=0.3016$ to $\bar\Gamma_{\mathrm{PRIME}}=0.1981$, corresponding to a relative reduction of $R_{\Gamma}=1-0.1981/0.3016=34.3\%$.

Gradient alignment increases more for semantic subgroups than for matched random groups. The reduction in $\Gamma$ therefore captures selective convergence of the update directions that exhibit the strongest competition, matching the optimization behavior in Equation~\ref{eq:3}. The matched non-conditional control next separates input conditioning from residual capacity.

We next construct a parameter-matched, non-conditional Dense residual to test whether any residual expansion or altered training trajectory would reduce $\Gamma$. The control uses the same Dense probability anchor and zero-residual initialization as PRIME, but all examples share the same added parameters and there is no input-dependent routing. Averaged over four architectures, three training seeds, and four semantic fields, the mean competition gaps are $\bar\Gamma_{\mathrm{Dense}}=0.3016$, $\bar\Gamma_{\mathrm{NonCond}}=0.2246$, and $\bar\Gamma_{\mathrm{PRIME}}=0.1981$.

The non-conditional residual reduces the gap by 0.0770 relative to Dense, showing that function-preserving residual capacity alleviates part of the shared compromise. PRIME provides a further reduction of 0.0265. Under the same budget and initialization, this additional reduction establishes a contribution from input-conditioned parameter organization beyond residual capacity and links PRIME's predictive gains to stronger optimization agreement.

\subsection{Capacity-Matched Controls and the Contribution of Input Conditioning}

\label{sec:capacity-controls}

This section answers RQ3 and evaluates the component-level design hierarchy in RQ4. We separate conditional organization from added capacity using two non-conditional Dense residual adapters. One approximately matches PRIME's added parameter count; the other approximately matches its added activation MACs. Both controls use the same zero-residual initialization and probability anchor as PRIME but contain no expert router. For budget-matched control $b$, define

\begin{equation}
\label{eq:29}
\mathcal A_b
=\mathrm{AUC}_{\mathrm{PRIME}}-
\mathrm{AUC}_{b}.
\end{equation}

The experiment uses the held-out Avazu test set, DNN, AutoInt, FiBiNET, DCNv2, and five paired seeds, yielding 20 paired runs for each comparison. These comparisons serve as diagnostic controls that isolate the effects of capacity, routing, anchoring, and aggregation. All variant configurations were fixed using validation data before held-out test evaluation, and no test result was used for subsequent tuning. Relative to the original Dense models, the parameter-matched and MAC-matched residuals improve mean AUC by 0.0019 and 0.0020, respectively. Non-conditional residual capacity is therefore useful. Both controls nevertheless remain below PRIME by 0.0012 and 0.0011 AUC. The resulting $\mathcal A_b$ measures the value of input-dependent expert organization after parameter or activation cost is matched.

\begin{figure}[t]
  \centering
  \includegraphics[width=\columnwidth]{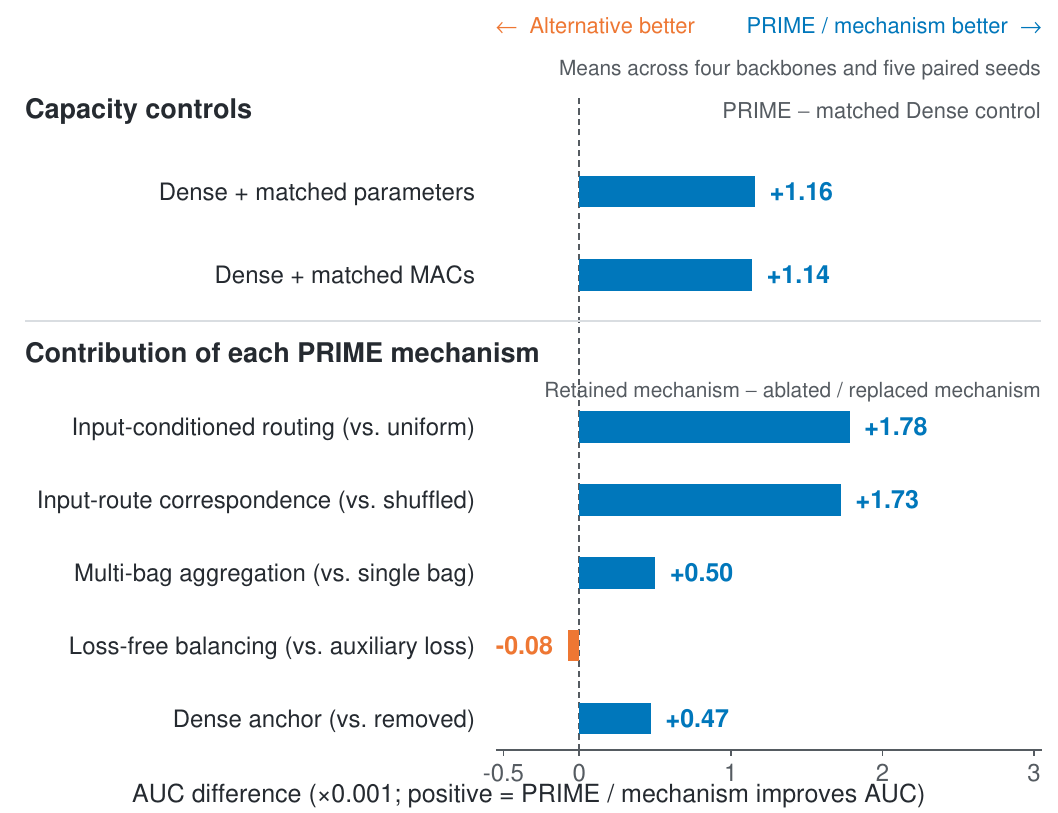}
  \caption{AUC differences between PRIME and capacity or mechanism controls. Bars report means over four backbones and five paired seeds. Every horizontal bar is oriented as PRIME, or the retained mechanism, minus the corresponding control or ablation. Positive values favor the PRIME design.}
  \Description{AUC advantage of PRIME over capacity-matched controls and mechanism alternatives on Avazu. The chart or architecture is interpreted quantitatively in the caption and surrounding text.}
  \label{fig:prime-4}
\end{figure}

Figure~\ref{fig:prime-4} reveals a clear design hierarchy. Uniform routing retains all expert parameters but removes input-dependent allocation. Permuted routing preserves aggregate load but breaks the input-expert correspondence. These interventions reduce mean AUC by 0.0018 and 0.0017, respectively, providing direct evidence for conditional specialization. Retaining the Dense anchor contributes 0.0005 mean AUC, confirming the function-preservation role of Equations~\ref{eq:6} and~\ref{eq:17}. Together with the parameter- and MAC-matched controls, these results establish function-preserving, input-conditioned parameter organization as the source of PRIME's advantage over ordinary capacity growth and unconditional ensembling. Multi-bag aggregation contributes a further 0.0005 mean AUC through multiple conditional estimates. Auxiliary-loss-free balancing matches the accuracy of loss-based balancing while preserving the single-task objective in Equation~\ref{eq:23} and eliminating an additional loss coefficient.

\subsection{Comparison with Conditional Parameterization and End-to-End Cost}

\label{sec:apg-efficiency}

This section completes RQ4 by comparing PRIME with the official FuxiCTR implementation of APG \cite{ref30}, a related input-conditioned parameterization method, and by measuring end-to-end cost. The comparison covers FiBiNET and DCNv2. For FiBiNET, APG is attached before bilinear interaction and retains the original Dense Top-NN. The methods use identical data splits, embedding dimensions, backbone hidden layers, optimizers, batch sizes, and early-stopping protocols. APG-specific hyperparameters are selected on the validation set without access to test data.

All methods are profiled on the same accelerator with batches of 4,096 examples, after 20 warm-up steps, using 100 timed inference steps and three independent repetitions.

The APG configurations train reliably on Avazu. Table~\ref{tab:prime-1} reports held-out test results over five seeds for the shared backbones. PRIME exceeds APG by 0.0030 AUC on FiBiNET and 0.0032 AUC on DCNv2, and all ten seed-level comparisons favor PRIME. On FiBiNET, APG improves over Dense, while PRIME provides a further gain with fewer parameters and lower latency; on DCNv2, PRIME exceeds both Dense and APG. Across both shared backbones, Dense-anchored conditional residuals are more consistent than direct generation of dynamic prediction parameters.

\begin{table}[t]
  \caption{Test accuracy and inference cost on Avazu. AUC and LogLoss are five-seed means; latency is the median per 4,096-example batch over three profiling repetitions. Bold accuracy values are best per backbone, while bold cost values mark the more efficient input-conditioned method.}
  \label{tab:prime-1}
  \centering
  \scriptsize
  \setlength{\tabcolsep}{3pt}
  \resizebox{\columnwidth}{!}{%
  \begin{tabular}{cccccc}
    \toprule
    Backbone & Method & Test AUC & Test LogLoss & Params (M) & Latency (ms) \\
    \midrule
    & Dense & 0.7600 & 0.3695 & 19.1582 & 2.1993 \\
    FiBiNET & APG & 0.7620 & 0.3684 & 19.4240 & 2.4191 \\
    & PRIME & \textbf{0.7650} & \textbf{0.3663} & \textbf{19.2784} & \textbf{2.3752} \\
    \cmidrule(lr){1-6}
    & Dense & 0.7626 & 0.3680 & 16.0000 & 1.1260 \\
    DCNv2 & APG & 0.7610 & 0.3689 & 17.0100 & 1.7270 \\
    & PRIME & \textbf{0.7642} & \textbf{0.3671} & \textbf{16.1200} & \textbf{1.4521} \\
    \bottomrule
  \end{tabular}%
  }
\end{table}

Table~\ref{tab:prime-1} shows that conditional capacity need not require a large deployment footprint. On FiBiNET, PRIME adds 0.6\% parameters and 8.0\% latency over Dense, while using 0.7\% fewer parameters and 1.8\% lower latency than APG. On DCNv2, the corresponding overheads over Dense are 0.8\% and 29.0\%, whereas the reductions relative to APG are 5.2\% and 15.9\%. PRIME therefore retains near-Dense parameter scale while improving both accuracy and efficiency over APG.

\subsection{Final Held-Out Results on Avazu and Criteo}

\label{sec:final-results}

To answer RQ5, both the PRIME configuration and all backbone settings are frozen after model selection on the validation set. Table~\ref{tab:final-results} reports the held-out test performance over five seeds for each architecture. \texttt{Dense} denotes the complete original backbone, and \texttt{PRIME} denotes the same backbone trained end to end with the plug-in enabled.

\begin{table*}[t]
  \caption{Architecture-level test results on Avazu and Criteo. Dense and PRIME entries are mean $\pm$ sample standard deviation over five paired seeds. $\Delta$ is PRIME$-$Dense for AUC and Dense$-$PRIME for LogLoss; positive values favor PRIME and are bold. The Mean row averages architecture-level means and within-architecture standard deviations over all 13 architectures; the paired-median row reports the median $\Delta$ over all 65 architecture--seed pairs.}
  \label{tab:final-results}
  \centering
  \footnotesize
  \setlength{\tabcolsep}{3pt}
  \resizebox{\textwidth}{!}{%
  \begin{tabular}{lcccccc@{\hspace{8pt}}cccccc}
    \toprule
    & \multicolumn{6}{c}{Avazu} & \multicolumn{6}{c}{Criteo} \\
    \cmidrule(lr){2-7}\cmidrule(lr){8-13}
    & \multicolumn{3}{c}{AUC} & \multicolumn{3}{c}{LogLoss}
    & \multicolumn{3}{c}{AUC} & \multicolumn{3}{c}{LogLoss} \\
    \cmidrule(lr){2-4}\cmidrule(lr){5-7}\cmidrule(lr){8-10}\cmidrule(lr){11-13}
    Backbone & Dense & PRIME & $\Delta$ & Dense & PRIME & $\Delta$
    & Dense & PRIME & $\Delta$ & Dense & PRIME & $\Delta$ \\
    \midrule
    AFM & $0.7402\pm0.0010$ & $0.7455\pm0.0004$ & \textbf{+0.0053} & $0.3812\pm0.0013$ & $0.3785\pm0.0008$ & \textbf{+0.0027} & $0.7711\pm0.0135$ & $0.7859\pm0.0025$ & \textbf{+0.0148} & $0.7623\pm0.5594$ & $0.4688\pm0.0045$ & \textbf{+0.2935} \\
    AutoInt & $0.7599\pm0.0048$ & $0.7646\pm0.0008$ & \textbf{+0.0047} & $0.3695\pm0.0023$ & $0.3668\pm0.0004$ & \textbf{+0.0027} & $0.6740\pm0.0306$ & $0.7944\pm0.0018$ & \textbf{+0.1204} & $2.2000\pm1.4766$ & $0.4557\pm0.0015$ & \textbf{+1.7442} \\
    DCN & $0.7641\pm0.0003$ & $0.7645\pm0.0007$ & \textbf{+0.0004} & $0.3671\pm0.0003$ & $0.3670\pm0.0003$ & \textbf{+0.0001} & $0.7799\pm0.0091$ & $0.8033\pm0.0010$ & \textbf{+0.0234} & $0.4682\pm0.0072$ & $0.4478\pm0.0009$ & \textbf{+0.0203} \\
    DNN & $0.7630\pm0.0001$ & $0.7641\pm0.0013$ & \textbf{+0.0011} & $0.3675\pm0.0001$ & $0.3668\pm0.0007$ & \textbf{+0.0007} & $0.7985\pm0.0042$ & $0.8019\pm0.0018$ & \textbf{+0.0034} & $0.4579\pm0.0054$ & $0.4504\pm0.0016$ & \textbf{+0.0075} \\
    DeepFM & $0.7636\pm0.0015$ & $0.7630\pm0.0022$ & -0.0006 & $0.3672\pm0.0008$ & $0.3678\pm0.0011$ & -0.0005 & $0.7048\pm0.0411$ & $0.7927\pm0.0015$ & \textbf{+0.0879} & $4.0806\pm4.2571$ & $0.4582\pm0.0017$ & \textbf{+3.6224} \\
    FiBiNET & $0.7600\pm0.0037$ & $0.7650\pm0.0007$ & \textbf{+0.0050} & $0.3695\pm0.0017$ & $0.3663\pm0.0004$ & \textbf{+0.0031} & $0.7969\pm0.0010$ & $0.8038\pm0.0007$ & \textbf{+0.0069} & $0.4535\pm0.0009$ & $0.4473\pm0.0006$ & \textbf{+0.0062} \\
    FwFM & $0.7488\pm0.0012$ & $0.7459\pm0.0015$ & -0.0030 & $0.3772\pm0.0009$ & $0.3782\pm0.0008$ & -0.0010 & $0.7657\pm0.0027$ & $0.7820\pm0.0033$ & \textbf{+0.0163} & $0.5288\pm0.0204$ & $0.4707\pm0.0035$ & \textbf{+0.0580} \\
    GDCN & $0.7620\pm0.0014$ & $0.7629\pm0.0014$ & \textbf{+0.0009} & $0.3685\pm0.0007$ & $0.3679\pm0.0008$ & \textbf{+0.0006} & $0.8066\pm0.0020$ & $0.8041\pm0.0061$ & -0.0025 & $0.4447\pm0.0020$ & $0.4469\pm0.0054$ & -0.0022 \\
    MaskNet & $0.7601\pm0.0002$ & $0.7619\pm0.0022$ & \textbf{+0.0018} & $0.3693\pm0.0004$ & $0.3685\pm0.0010$ & \textbf{+0.0008} & $0.8100\pm0.0004$ & $0.8099\pm0.0013$ & -0.0001 & $0.4416\pm0.0004$ & $0.4417\pm0.0012$ & -0.0001 \\
    NFM & $0.7568\pm0.0009$ & $0.7644\pm0.0004$ & \textbf{+0.0076} & $0.3702\pm0.0004$ & $0.3669\pm0.0004$ & \textbf{+0.0033} & $0.7768\pm0.0373$ & $0.7943\pm0.0020$ & \textbf{+0.0175} & $0.4993\pm0.0948$ & $0.4605\pm0.0070$ & \textbf{+0.0389} \\
    PNN & $0.7600\pm0.0042$ & $0.7634\pm0.0013$ & \textbf{+0.0034} & $0.3692\pm0.0022$ & $0.3673\pm0.0007$ & \textbf{+0.0018} & $0.8037\pm0.0011$ & $0.8057\pm0.0003$ & \textbf{+0.0020} & $0.4473\pm0.0010$ & $0.4454\pm0.0003$ & \textbf{+0.0019} \\
    Wide\&Deep & $0.7632\pm0.0024$ & $0.7643\pm0.0002$ & \textbf{+0.0011} & $0.3675\pm0.0009$ & $0.3670\pm0.0001$ & \textbf{+0.0005} & $0.7636\pm0.0266$ & $0.7960\pm0.0004$ & \textbf{+0.0324} & $0.4813\pm0.0206$ & $0.4545\pm0.0004$ & \textbf{+0.0268} \\
    DCNv2 & $0.7626\pm0.0015$ & $0.7642\pm0.0013$ & \textbf{+0.0016} & $0.3680\pm0.0010$ & $0.3671\pm0.0008$ & \textbf{+0.0009} & $0.7977\pm0.0055$ & $0.8031\pm0.0027$ & \textbf{+0.0054} & $0.4530\pm0.0050$ & $0.4482\pm0.0026$ & \textbf{+0.0048} \\
    \midrule
    Mean & $0.7588\pm0.0018$ & $0.7611\pm0.0011$ & \textbf{+0.0023} & $0.3701\pm0.0010$ & $0.3689\pm0.0007$ & \textbf{+0.0012} & $0.7730\pm0.0135$ & $0.7982\pm0.0020$ & \textbf{+0.0252} & $0.9014\pm0.4962$ & $0.4535\pm0.0024$ & \textbf{+0.4479} \\
    Paired median $\Delta$ & -- & -- & \textbf{+0.0022} & -- & -- & \textbf{+0.0011} & -- & -- & \textbf{+0.0066} & -- & -- & \textbf{+0.0081} \\
    \bottomrule
  \end{tabular}
  }
\end{table*}

On Avazu, PRIME raises macro-average AUC from 0.7588 to 0.7611 and lowers LogLoss from 0.3701 to 0.3689, with a median paired $\Delta\mathrm{AUC}$ of 0.0022. When the five-seed standard deviation is averaged across architectures, PRIME reduces AUC dispersion from 0.0018 to 0.0011 and LogLoss dispersion from 0.0010 to 0.0007. These simultaneous improvements in predictive quality and average cross-seed dispersion hold across a diverse suite of plain DNNs, explicit crosses, attention, bilinear interactions, factorization-based models, and mask-based networks.

\begin{sloppypar}
On Criteo, every Dense/PRIME pair uses identical configurations for the shared backbone and training protocol; enabling PRIME is the only structural change. Under this frozen common protocol, Dense AutoInt and DeepFM repeatedly suffer severe optimization degradation, whereas their PRIME counterparts remain stable. We retain these runs and include them in the macro-average to preserve the uniform evaluation standard. Consequently, the aggregate gains of +0.0252 AUC and +0.4479 LogLoss summarize the complete protocol but combine conventional predictive improvements with recovery from Dense optimization failure. Cross-architecture effectiveness is established separately by the positive mean $\Delta\mathrm{AUC}$ on 11 of 13 architectures and the +0.0066 median over all paired runs, both of which are substantially less sensitive to the two degraded Dense baselines.
\end{sloppypar}

\section{Discussion}

\label{sec:discussion}

\subsection{How Conditional Residual Experts Reduce Shared Top-NN Competition}

\label{sec:discussion-parameterization}

PRIME changes where heterogeneous examples are allowed to disagree. In a Dense baseline, every example updates the same late-stage coordinates, forcing subgroup-specific signals into a single shared direction. PRIME instead decomposes the decision into a shared Dense prediction and an input-conditioned residual. Common regularities remain in the anchor, while input-specific deviations are absorbed by selected low-rank experts. It therefore reorganizes parameter sharing without subgroup labels or replicated prediction networks.

The 34.3\% reduction in the semantic subgroup competition gap provides direct optimization evidence for this mechanism. Parameter- and MAC-matched Dense controls remain below PRIME, a non-conditional residual recovers only part of the gain, and uniform or permuted routing removes the remaining advantage. The improvement consequently requires both residual capacity and input-conditioned organization. Routed experts act as local correction subspaces around a common function, making updates at the shared path more compatible.

To test whether function preservation matters, we compared PRIME with two development-stage alternatives that modify the Dense path: direct hidden-layer subspace replacement and multi-layer bagged replacement. Across four Avazu architectures and five seeds, a win required simultaneous validation AUC and LogLoss improvements over Dense. PRIME won 15 of 20 pairs, versus 4 of 20 and 8 of 20 for the two alternatives. Zero-residual initialization and bounded mixing preserve a usable predictor while specialization develops; direct replacement instead changes the baseline immediately and propagates early expert errors.

\fixedsubsection{When PRIME Works: Aligning Experts with the Shared Decision Path}

\label{sec:discussion-boundaries}

Cross-architecture results identify one attachment principle: the router must observe the representation consumed by the shared decision mapping, and the residual must correct that same stage. This alignment, rather than the interaction operator itself, separates consistently positive backbones from the boundary cases.

FwFM and GDCN expose the boundary because their improvement directions reverse across datasets. FwFM has no shared MLP Top-NN; it predicts from linear and field-weighted pairwise terms \cite{ref12}. PRIME raises its mean AUC by 0.0163 on Criteo but reduces it by 0.0030 on Avazu. Here the plug-in acts as an auxiliary conditional predictor: extra nonlinear capacity can benefit one distribution, but it does not systematically reorganize a contested shared mapping.

The subgroup diagnostic explains why attachment location matters. The competition gap $\Gamma$ is defined from gradients with respect to a particular shared mapping; it can be reduced systematically only when routing exposes the subgroup distinctions driving those gradients. GDCN improves by 0.0009 AUC on Avazu but declines by 0.0025 on Criteo. Its decision uses a gated crossed representation \cite{ref25}, whereas PRIME currently routes on pre-cross embeddings. The controls identify pre-cross router conditioning as a major contributor to GDCN's degradation. An additional five-seed Criteo experiment in Appendix~\ref{sec:appendix-gdcn-controls} supports this interpretation: moving the router to an intermediate cross layer raises mean AUC from 0.8041 to 0.8063. When subgroup distinctions emerge after crossing, experts are assigned before the competing directions become visible. They may still learn useful correlations, but their specialization is no longer coupled to GDCN's final decision path.

Together, these cases motivate architecture-aware attachment: routing should use decision-relevant representations, and the residual should target an explicit shared prediction path.

\section{Conclusion}

\label{sec:conclusion}

This paper identifies a parameter-sharing constraint in fully shared CTR top networks. On Avazu, semantic subgroups consistently produce lower Top-NN gradient alignment than size- and label-matched random groups. Heterogeneous examples therefore impose less compatible update directions on a single shared mapping, motivating conditional parameter organization within the decision layer.

PRIME addresses this constraint through function-preserving conditional adaptation. The Dense prediction remains anchored, and zero-residual initialization preserves the initial function. Input-dependent routing combines bounded low-rank logit corrections without replicating the backbone interaction module. Multi-bag aggregation supplies complementary conditional estimates, while EMA load adjustment limits expert imbalance. This design introduces specialization while retaining the optimization stability and statistical strength of the shared path.

Across held-out Avazu and Criteo tests, PRIME improves mean AUC for 11 of 13 architectures on each dataset. On the FiBiNET and DCNv2 backbones shared with APG, PRIME wins all ten Avazu seed-level comparisons. It also remains above parameter- and MAC-matched Dense controls, while routing controls show that the gain arises from input-conditioned expert organization rather than capacity alone. After PRIME training, the semantic subgroup competition gap decreases by 34.3\%, connecting stronger predictions with more compatible shared updates. Overall, PRIME provides a compact, function-preserving, and budget-controlled route to conditional adaptation of shared CTR Top-NNs across heterogeneous backbones.

\section*{Ethical Considerations}

This study uses public, de-identified CTR benchmark data and involves no human recruitment or personally identifiable information. % CTR systems nevertheless influence exposure to items, information, and commercial opportunities; without safeguards, improved prediction may amplify historical bias or unequal treatment. PRIME changes model capacity and parameter sharing but does not correct biased labels, unfair objectives, privacy leakage, or miscalibration under distribution shift. Production deployment should retain platform privacy and data-governance controls and monitor subgroup calibration, exposure outcomes, router utilization, and performance drift.

\clearpage
\appendix
\renewcommand{\thetable}{S\arabic{table}}
\renewcommand{\theHtable}{supp.\arabic{table}}
\setcounter{table}{0}
\twocolumn[
  \begin{center}
    {\Large\bfseries Supplementary Material}
  \end{center}
  \vspace{0.35\baselineskip}
]

\section{Reproducibility and Design Map}

All main experiments use Adam with learning rate $10^{-3}$, batch size 4,096, and validation AUC for checkpoint selection. Training proceeds for at most 100 epochs, stops after two consecutive validation checks without improvement, and restores the best checkpoint. The embedding dimension is 10. A typical Top-NN has three hidden layers of width 400; dropout, batch normalization, and regularization follow the public configuration of each backbone and remain fixed within every Dense/PRIME pair.

The frozen PRIME configuration uses $G=4$ bags, $M=8$ experts per bag, rank $q=16$, Dense weight $\alpha=0.5$, routing temperature $T=1$, moving-load coefficient $\mu=0.99$, bias step size $\eta_b=10^{-3}$, and correction bound $c_{\max}=2$. The software stack consists of Python 3.10.16, PyTorch 2.6.0, CUDA 12.6, cuDNN 8.9.5, and FuxiCTR 2.3.10. Training uses 16 PPU-ZW810E accelerators. Checkpoints, resolved configurations, validation records, and test records are stored separately, while hashes of the feature map, configuration, and code revision permit run-level verification. No job is repeated because of an unfavorable metric.

The design hypotheses and their corresponding tests are summarized below.

\begin{enumerate}
  \item \textbf{Conditional correction.} Weak alignment across semantic subgroups motivates input-conditioned routing. Uniform and permuted-routing controls test whether preserving the input--expert correspondence matters.
  \item \textbf{Function preservation.} Perturbing a validated prediction path motivates the Dense anchor and zero-residual initialization. The no-anchor ablation tests this requirement.
  \item \textbf{Budget control.} Replicating complete Top-NNs is unnecessary for residual correction. Parameter- and MAC-matched non-conditional adapters test whether low-rank conditional organization contributes beyond capacity alone.
  \item \textbf{Conditional aggregation.} Multiple expert bags provide several conditional estimates. The single-bag control tests whether their average supplies an additional empirical benefit.
  \item \textbf{Objective preservation.} EMA load bias separates load adjustment from CTR gradients. The auxiliary-loss control tests whether comparable accuracy can be obtained without introducing another loss coefficient.
\end{enumerate}

The first three items define PRIME's core parameterization and efficiency constraints; the final two provide complementary stabilization and aggregation.

\section{Complete APG Comparison}

Table~\ref{tab:supp-apg-complete} reports the complete Avazu comparison with APG on FiBiNET and DCNv2. For FiBiNET, APG applies an input-conditioned low-rank transform to field embeddings before bilinear interaction while retaining the original bilinear module and Dense Top-NN; DCNv2 uses self-wise APG parameter generation in its prediction path. All configurations use the same data split, embedding dimension, optimizer, batch size, early-stopping protocol, and five paired random seeds.

\begin{table*}[t]
  \caption{Complete held-out Avazu comparison with APG. AUC and LogLoss are five-seed means; latency is the median per 4,096-example batch over three profiling repetitions. Seed columns report held-out AUC. $\Delta$ is PRIME$-$APG; positive AUC and negative LogLoss or cost differences favor PRIME. Bold values mark improvements over APG.}
  \label{tab:supp-apg-complete}
  \centering
  \fontsize{7.45pt}{8.8pt}\selectfont
  \begin{tabular*}{\textwidth}{@{\extracolsep{\fill}}llrrrrrrrrr@{}}
    \toprule
    & & \multicolumn{2}{c}{Five-seed mean} & \multicolumn{2}{c}{Deployment} & \multicolumn{5}{c}{Held-out AUC by seed} \\
    \cmidrule(lr){3-4}\cmidrule(lr){5-6}\cmidrule(lr){7-11}
    Backbone & Method & AUC & LogLoss & Params (M) & Latency (ms) & 2021 & 190034 & 27011 & 948432 & 992817 \\
    \midrule
    FiBiNET & Dense & 0.7600 & 0.3695 & 19.1582 & 2.1993 & -- & -- & -- & -- & -- \\
             & APG & 0.7620 & 0.3684 & 19.4240 & 2.4191 & 0.7619 & 0.7641 & 0.7616 & 0.7619 & 0.7606 \\
             & PRIME & \textbf{0.7650} & \textbf{0.3663} & \textbf{19.2784} & \textbf{2.3752} & \textbf{0.7648} & \textbf{0.7644} & \textbf{0.7657} & \textbf{0.7643} & \textbf{0.7657} \\
             & $\Delta$ & \textbf{+0.0030} & \textbf{$-$0.0021} & \textbf{$-$0.1456} & \textbf{$-$0.0439} & \textbf{+0.0030} & \textbf{+0.0002} & \textbf{+0.0041} & \textbf{+0.0024} & \textbf{+0.0051} \\
    \midrule
    DCNv2 & Dense & 0.7626 & 0.3680 & 16.0000 & 1.1260 & -- & -- & -- & -- & -- \\
          & APG & 0.7610 & 0.3689 & 17.0100 & 1.7270 & 0.7609 & 0.7601 & 0.7611 & 0.7621 & 0.7609 \\
          & PRIME & \textbf{0.7642} & \textbf{0.3671} & \textbf{16.1200} & \textbf{1.4521} & \textbf{0.7622} & \textbf{0.7642} & \textbf{0.7659} & \textbf{0.7641} & \textbf{0.7645} \\
          & $\Delta$ & \textbf{+0.0032} & \textbf{$-$0.0018} & \textbf{$-$0.8900} & \textbf{$-$0.2749} & \textbf{+0.0013} & \textbf{+0.0041} & \textbf{+0.0048} & \textbf{+0.0020} & \textbf{+0.0036} \\
    \bottomrule
  \end{tabular*}
\end{table*}

PRIME achieves higher AUC in all ten paired seed-level comparisons; the smallest margin is +0.0002 on FiBiNET at seed 190034. On FiBiNET, APG improves AUC over Dense by 0.0020, while PRIME adds a further 0.0030 and reduces LogLoss by another 0.0021 relative to APG. On DCNv2, PRIME exceeds both Dense and APG in predictive accuracy. Across both backbones, PRIME also uses fewer parameters and has lower inference latency than APG.

\section{Routing Diagnostics and Dense-Weight Sensitivity}

Across 16 Avazu runs with complete diagnostic records, normalized routing entropy over eight experts per bag is $0.432\pm0.085$, corresponding to $2.49\pm0.47$ effective experts. The maximum mean load of one expert is $0.425\pm0.116$, above the uniform value of 0.125. Multiple experts remain active while preserving pronounced specialization. The observed load profile is consistent with the EMA load-bias update preventing single-expert collapse without erasing the expert preferences required for conditional routing.

With development seed 190034 and four architectures, $\alpha=0.25$, $0.5$, and $0.75$ produce mean validation AUCs of 0.7493, 0.7480, and 0.7474, with LogLoss values of 0.3947, 0.3955, and 0.3959. We then compare 0.25 and the frozen default 0.5 across four architectures and five seeds. Their mean AUCs are 0.7483 and 0.7480, with closely matched LogLoss. DNN and FiBiNET favor 0.5, whereas AutoInt and DCNv2 favor 0.25. These complementary directions support retaining the prespecified $\alpha=0.5$ as a balanced setting across backbones.

\section{GDCN Attachment Controls on Criteo}
\label{sec:appendix-gdcn-controls}

To examine why the original PRIME configuration trails GDCN on Criteo, we keep the GDCN backbone and PRIME capacity fixed while changing the representations supplied to the router and experts. Tables~\ref{tab:supp-gdcn-seeds} and~\ref{tab:supp-gdcn-controls} report the seed-level AUCs and complete five-seed attachment controls, respectively. \emph{Pre} denotes flattened embeddings before the gated cross network, \emph{Cross-1} and \emph{Cross-2} denote the outputs of its first and second cross layers, and \emph{Post} denotes the final crossed representation. Unless stated otherwise, the controls use input LayerNorm, an attached router gradient, and the bounded probability-mixture output. All results are held-out test metrics over the same five paired seeds.

\begin{table}[H]
  \caption{Seed-level AUC for GDCN/Criteo router controls, with means computed from unrounded run-level metrics. Experts remain on pre-cross embeddings in the cross-layer variants.}
  \label{tab:supp-gdcn-seeds}
  \centering
  \scriptsize
  \setlength{\tabcolsep}{2.5pt}
  \resizebox{\columnwidth}{!}{%
  \begin{tabular}{rrrrrr}
    \toprule
    Seed & Dense & Original & Cross-1 & Cross-2 & Cross-2, no LN \\
    \midrule
    2021   & 0.8081 & 0.8088 & 0.8056 & 0.8046 & 0.8046 \\
    190034 & 0.8031 & 0.8010 & 0.8070 & 0.8081 & 0.8084 \\
    27011  & 0.8078 & 0.8071 & 0.8054 & 0.8071 & 0.8073 \\
    948432 & 0.8070 & 0.8090 & 0.8063 & 0.8082 & 0.8061 \\
    992817 & 0.8073 & 0.7949 & 0.8073 & 0.8033 & 0.8066 \\
    \midrule
    Mean   & 0.8066 & 0.8041 & 0.8063 & 0.8063 & 0.8066 \\
    \bottomrule
  \end{tabular}%
  }
\end{table}

\begin{table}[H]
  \caption{Complete five-seed GDCN/Criteo attachment controls. Entropy is the unnormalized mean router entropy in nats over eight experts per bag; max load is the largest mean expert probability. The logit row replaces bounded probability mixing with a direct logit residual.}
  \label{tab:supp-gdcn-controls}
  \centering
  \scriptsize
  \setlength{\tabcolsep}{2.5pt}
  \resizebox{\columnwidth}{!}{%
  \begin{tabular}{lllrrrr}
    \toprule
    Configuration & Router & Expert & AUC & LogLoss & Entropy & Max load \\
    \midrule
    Dense & -- & -- & 0.8066 & 0.4447 & -- & -- \\
    Original PRIME & Pre & Pre & 0.8041 & 0.4469 & -- & -- \\
    Post/Post & Post & Post & 0.8059 & 0.4452 & 0.9674 & 0.5328 \\
    Post/Pre & Post & Pre & 0.8061 & 0.4452 & 0.9843 & 0.5418 \\
    Pre/Post & Pre & Post & 0.8035 & 0.4473 & 0.4086 & 0.9021 \\
    Post/Pre, detached & Post & Pre & 0.8051 & 0.4460 & 0.8986 & 0.6563 \\
    Post/Pre, logit & Post & Pre & 0.7975 & 0.4529 & 0.9375 & 0.4424 \\
    Cross-1/Pre & Cross-1 & Pre & 0.8063 & 0.4450 & 0.9273 & 0.5928 \\
    Cross-2/Pre & Cross-2 & Pre & 0.8063 & 0.4451 & 0.9817 & 0.5672 \\
    Cross-2/Pre, no LN & Cross-2 & Pre & 0.8066 & 0.4448 & 1.1695 & 0.5302 \\
    \bottomrule
  \end{tabular}%
  }
\end{table}

Moving the router from pre-cross embeddings to an intermediate cross layer raises mean AUC from 0.8041 to 0.8063. The more diagnostic reversal holds the expert source fixed: Post/Pre reaches 0.8061 AUC, whereas Pre/Post falls to 0.8035. The latter also has substantially lower router entropy (0.4086 versus 0.9843) and a higher maximum expert load (0.9021 versus 0.5418). These controls identify pre-cross router conditioning, rather than insufficient expert depth, as a major contributor to the original degradation.

The remaining variants delimit the mechanism. Post/Pre is slightly stronger than Post/Post, so exact coincidence between router and expert inputs is unnecessary. Detaching the router gradient lowers mean AUC, while direct logit residuals produce the weakest result. GDCN therefore benefits from routing on decision-relevant crossed representations while retaining PRIME's Dense anchor and bounded probability mixture.

\section{End-to-End Profiling Details}

Each method is profiled on the same accelerator with batch size 4,096, warmed up for 20 steps, timed for 100 inference steps and 50 training steps, and repeated independently three times. A training step includes forward propagation, backpropagation, and an optimizer update. Inference uses \texttt{model.eval()} without gradient tracking. PRIME's deployment path omits balancing losses, orthogonality Gram matrices, and host-side routing diagnostics used only during training or analysis. Table~\ref{tab:prime-efficiency} reports median absolute latency and the median paired ratio to the corresponding Dense backbone within each repetition.

\begin{table*}[t]
  \caption{End-to-end profiling on Avazu. Absolute latencies are medians of three independent repetitions, and ratios use the matching Dense backbone. Each inference batch and training step processes 4,096 examples. Bold values mark the lower cost between APG and PRIME for a given backbone.}
  \label{tab:prime-efficiency}
  \centering
  \fontsize{7.45pt}{8.8pt}\selectfont
  \begin{tabular*}{\textwidth}{@{\extracolsep{\fill}}llrrrrrr@{}}
    \toprule
    & & & \multicolumn{2}{c}{Inference} & \multicolumn{2}{c}{Training} & \\
    \cmidrule(lr){4-5}\cmidrule(lr){6-7}
    Backbone & Method & Parameter ratio & Latency (ms) & Ratio & Latency (ms) & Ratio & Peak-memory ratio \\
    \midrule
    FiBiNET & Dense & 1.0000 & 2.1993 & 1.0000 & 14.9712 & 1.0000 & 1.0000 \\
             & APG & 1.0139 & 2.4191 & 1.1001 & \textbf{15.7662} & \textbf{1.0535} & 1.5712 \\
             & PRIME & \textbf{1.0063} & \textbf{2.3752} & \textbf{1.0849} & 16.1865 & 1.0840 & \textbf{1.0062} \\
    \midrule
    DCNv2 & Dense & 1.0000 & 1.1260 & 1.0000 & 8.8583 & 1.0000 & 1.0000 \\
          & APG & 1.0632 & 1.7270 & 1.5337 & 10.9516 & 1.2402 & 1.8738 \\
          & PRIME & \textbf{1.0075} & \textbf{1.4521} & \textbf{1.2896} & \textbf{10.3558} & \textbf{1.1690} & \textbf{1.0231} \\
    \bottomrule
  \end{tabular*}
\end{table*}

Across the two backbones, PRIME adds only 0.6\%--0.8\% parameters over Dense. Its training-step overhead ranges from 8.4\% on FiBiNET to 16.9\% on DCNv2. PRIME is faster and substantially more memory-efficient than APG on DCNv2. On FiBiNET, pre-bilinear APG is 2.6\% faster per training step, whereas PRIME lowers peak training memory by 36.0\% and remains faster at inference. Its soft router evaluates all low-rank experts in every bag, so the measured latency reflects full conditional estimation \cite{ref41}. Sparse scheduling and operator fusion offer direct paths to further acceleration.

\section{Development-Stage Structural Controls}

During development, we evaluated two expert designs that modify the Dense path. The first replaces hidden-layer linear transformations with subspace experts. The second inserts bagged experts at multiple hidden stages. These alternatives provide structural controls for the function-preservation requirement. Because they alter the baseline mapping, early expert errors propagate through subsequent nonlinear layers and their optimization trajectories begin from a different function than the original model.

Across four Avazu architectures and five seeds, PRIME improves both validation AUC and LogLoss in 15 of 20 pairs, compared with 4 of 20 for direct hidden-layer subspace replacement and 8 of 20 for the best multi-layer bagged replacement. A win requires simultaneous improvement in both metrics. This decisive margin isolates function preservation as a central requirement and motivates PRIME's Dense-anchored design.

\end{document}